\documentclass[10pt,twocolumn,letterpaper]{article}
\usepackage[pagenumbers]{cvpr} 
\usepackage{multirow}
\usepackage{colortbl}
\usepackage{xcolor}
\usepackage{pifont}
\usepackage{makecell}

\usepackage{pythonhighlight}
\usepackage{marvosym}
\usepackage[linesnumbered,ruled,vlined]{algorithm2e}

\definecolor{cvprblue}{rgb}{0.21,0.49,0.74}
\usepackage[pagebackref,breaklinks,colorlinks,allcolors=cvprblue]{hyperref}
\usepackage[accsupp]{axessibility}

\title{The 1st Solution for 4th PVUW MeViS Challenge: Unleashing the Potential of Large Multimodal Models for Referring Video Segmentation}
\author{Hao Fang, Runmin Cong, Xiankai Lu, Zhiyang Chen, Wei Zhang\\
Shandong University\\
Team: MVP-Lab
}

\begin{document}
\maketitle

\begin{abstract}
Motion expression video segmentation is designed to segment objects in accordance with the input motion expressions. In contrast to the conventional Referring Video Object Segmentation (RVOS), it places emphasis on motion as well as multi-object expressions, making it more arduous. Recently, Large Multimodal Models (LMMs) have begun to shine in RVOS due to their powerful vision-language perception capabilities. In this work, we propose a simple and effective inference optimization method to fully unleash the potential of LMMs in referring video segmentation. Firstly, we use Sa2VA as our baseline, which is a unified LMM for dense grounded understanding of both images and videos. Secondly, we uniformly sample the video frames during the inference process to enhance the model's understanding of the entire video. Finally, we integrate the results of multiple expert models to mitigate the erroneous predictions of a single model. Our solution achieved 61.98\% $\mathcal{J} \& \mathcal{F}$ on the MeViS test set and ranked 1st place in the 4th PVUW Challenge MeViS Track at CVPR 2025.
\end{abstract}

\section{Introduction}
\label{sec:intro}
Referring video object segmentation (RVOS) is a continually evolving task that aims to segment target objects in video, referred to by linguistic expressions. Recently, a new large-scale dataset called Motion expressions Video Segmentation (MeViS)~\cite{ding2023mevis} has been proposed, which focuses on segmenting objects in video content based on a sentence describing the motion of the objects. Compared to traditional RVOS datasets, it emphasizes motion and multi-object expression, making it more challenging.

Most early RVOS approaches~\cite{khoreva2019video,liang2021rethinking} adopt multi-stage and complex pipelines that take the bottom-up or top-down paradigms to segment each frame separately. Meanwhile, compared to rely on complicated pipelines, MTTR~\cite{botach2022end} and Referformer~\cite{wu2022language} first adopt end-to-end framework modeling the task as the a sequence prediction problem, which greatly simplifies the pipeline. Based on the end-to-end architecture of Transformer, SOC~\cite{luo2023soc} and MUTR~\cite{yan2024referred} achieve excellent performance by efficiently aggregating intra and inter-frame information. For example, the 1st and 2nd place solution for MeViS Track in 3th PVUW Workshop~\cite{ding2024pvuw} both involve fine-tuning MUTR~\cite{yan2024referred} on MeViS~\cite{ding2023mevis}. 
LMPM~\cite{ding2023mevis} and DsHmp~\cite{he2024decoupling} add a motion expression understanding module to the video instance segmentation~\cite{heo2022vita,fang2024learning,fang2024unified} framework, specifically designed for processing motion expression video segmentation.
To fully utilize the capabilities of existing video segmentation models, the 1st place solution~\cite{fang2024uninext} for 6th LSVOS Challenge RVOS Track~\cite{ding2024lsvos} integrate strengths of that leading RVOS and VOS models~\cite{yan2023universal,cheng2024putting,pan2024video} to build up a “refine after fine-tuning” pipeline for motion expression video segmentation.

Thanks to the achievements of Large Language Models (LLMs), Large Multimodal Models (LMMs) have seen unprecedented development. Recent studies~\cite{lai2024lisa,ren2024pixellm} have explored the implementation of LMMs to produce object masks in a novel reasoning segmentation task, which enhances the applicability to real-world applications. Inspired by this, VISA~\cite{yan2024visa} and ViLLa~\cite{zheng2024villa} introduce a new task Reasoning Video Object Segmentation, which aims to segment and track objects in videos given implicit texts. They collect large-scale datasets and propose reasoning video segmentation models based on LMMs. Sa2VA~\cite{yuan2025sa2va} is the first unified model for dense grounded understanding of both images and videos. Sa2VA combines SAM 2~\cite{ravi2024sam}, a foundation video segmentation model, with LLaVA~\cite{liu2023visual}, an advanced vision-language model, and unifies text, image, and video into a shared LLM token space. Experiments show that Sa2VA achieves state-of-the-art across multiple tasks, particularly in RVOS, highlighting its potential for motion expression video segmentation. 

\begin{figure*}
  \centering
  \includegraphics[width=1.0\linewidth]{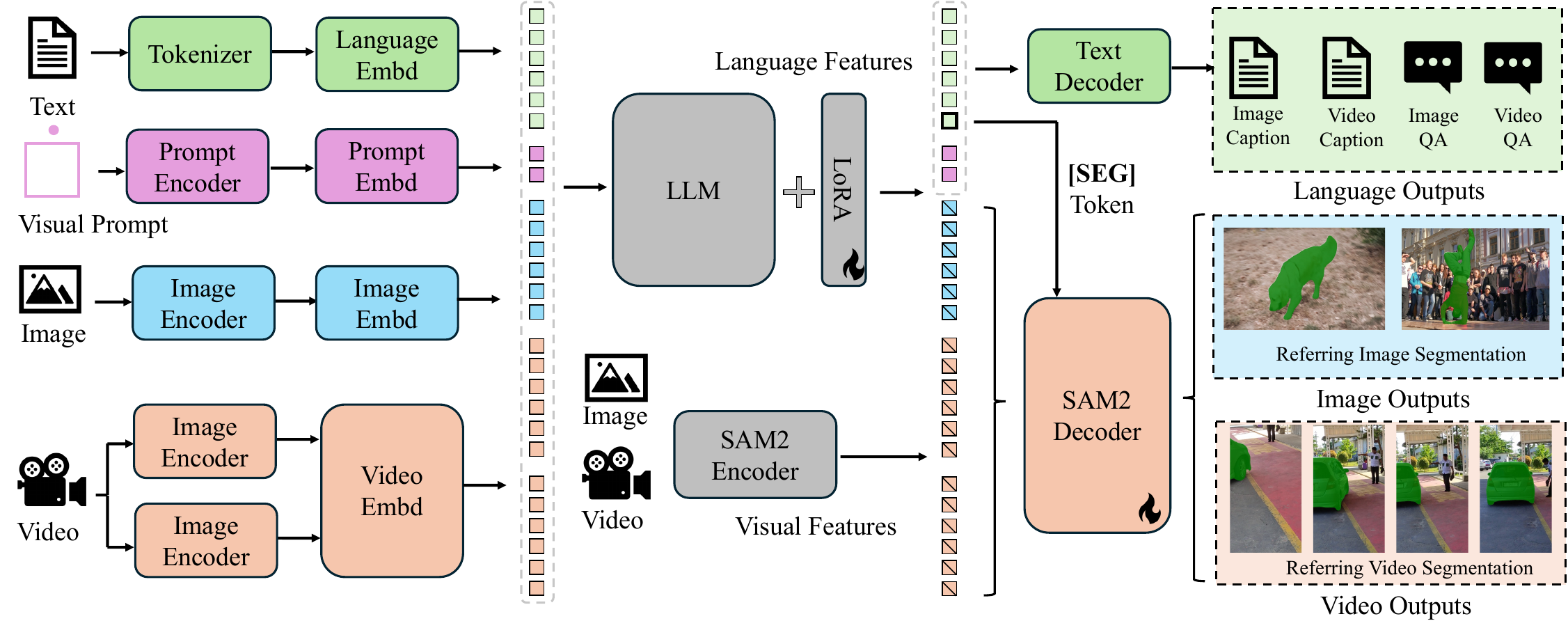}
  \caption{\textbf{The architecture of Sa2VA~\cite{yuan2025sa2va}.} The model first encodes the input texts, visual prompts, images, and videos into token embeddings. These tokens are then processed through a large language model (LLM). The output text tokens are used to generate the ``[SEG]'' token and associated language outputs. The SAM 2 decoder receives the image and video features from the SAM 2 encoder, along with the ``[SEG]'' token, to generate corresponding image and video masks.}
  \label{fig:method}
\end{figure*}

In this work, we propose a simple and effective inference optimization method to fully unleash the potential of LMMs in referring video segmentation. Firstly, we use Sa2VA~\cite{yuan2025sa2va} as our baseline, which uses LLM to generate instruction tokens that guide SAM 2 in producing precise masks, enabling a grounded, multi-modal understanding of both static and dynamic visual content. Secondly, during the inference process, Sa2VA defaults to inputting the first few frames into LLM, but MeViS is a long video dataset, which results in a significant loss of video information. We uniformly sample the video frames to enhance the model's understanding of the entire video. Finally, we find that Sa2VA does not necessarily perform better with a larger number of parameters and more sampling frames, as each configuration has its own strengths in different videos. And for some videos that cannot be accurately segmented by LMMs, the classic RVOS model may handle them very well. So we integrate the results of multiple expert models to mitigate the erroneous predictions of a single model. 

In this year, Pixel-level Video Understanding in the Wild Challenge (PVUW) challenge adds two new tracks, Complex Video Object Segmentation Track based on MOSE~\cite{ding2023mose} and Motion Expression guided Video Segmentation track based on MeViS~\cite{ding2023mevis}. In the two new tracks, additional videos and annotations that feature challenging elements are provided, such as the disappearance and reappearance of objects, inconspicuous small objects, heavy occlusions, and crowded environments in MOSE~\cite{ding2023mose}. Moreover, a new motion expression guided video segmentation dataset MeViS~\cite{ding2023mevis} is provided to study the natural language-guided video understanding in complex environments. These new videos, sentences, and annotations enable us to foster the development of a more comprehensive and robust pixel-level understanding of video scenes in complex environments and realistic scenarios.
Thanks to the superior performance of Sa2VA~\cite{yuan2025sa2va} and UNINEXT-Cutie~\cite{fang2024uninext}, our solution achieved 61.98\% $\mathcal{J} \& \mathcal{F}$ on the MeViS test set and ranked 1st place in the 4th PVUW Challenge MeViS Track at CVPR 2025.

\section{Method}
\label{sec:metho}
The input of RVOS contains a video sequence $\mathcal{S} = \left\{X_t\in \mathbb{R}^{3 \times H \times W} \right\}_{t=1}^N $ with $N$ frames and a corresponding referring expression $\mathcal{T} = \left\{ T_l \right\}_{l=1}^L $ with \textit{L} words. Our solution consists of three parts: Baseline(~\cref{sec:baseline}), Inference(~\cref{sec:inference}), and Aggregation(~\cref{sec:aggregation}).

\subsection{Baseline}
\label{sec:baseline}
We adopt Sa2VA~\cite{yuan2025sa2va} as our baseline to obtain mask sequences $\mathcal{M} = \{M_t\}_{t=1}^N$ that are correlated with language descriptions:
\begin{equation}
    \mathcal{M} = \mathcal{F}^{rvos}\left( \mathcal{S}, \mathcal{T}\right),
\end{equation}
where $\mathcal{F}^{rvos}$ denotes the Sa2VA model. The overall architecture of Sa2VA is shown in Fig.~\ref{fig:method}. It contains two parts: the LLaVA-like model and SAM 2. 

\noindent \textbf{Pre-trained LMMs.} Sa2VA adopts pre-trained LLaVA-like models as the LMMs. It contains one visual encoder, one visual projection layer, and one LLM. The visual encoder takes input images, video, and sub-images as inputs. The visual projection layer maps inputs into visual tokens. These tokens, combined with the input text tokens, are the input of LLMs and the LLMs generate the text token prediction based on them. Note that Sa2VA adopts pre-trained LMMs following previous works~\cite{lai2024lisa,yan2024visa} to leverage their strong capability. For both image and video chat datasets, it follows the same pipeline~\cite{wang2024qwen2} without modification.

\noindent \textbf{Decoupled Design.} Sa2VA append SAM 2 alongside the pre-trained LLaVA model. It does not take the SAM 2's output tokens (visual features or decoder outputs) into LLM. There are three reasons. First, Sa2VA makes the combination as simple as possible without increasing extra computation costs. Secondly, adding extra tokens needs an extra alignment process. Thirdly, via this design, it can fully make our work as a plug-in-play framework to utilize pre-trained LMMs since the LMM community goes fast. Thus, Sa2VA adopts a decoupled design without introducing further communication between LLaVA and SAM 2.

\noindent \textbf{Tuning SAM 2 Decoder via SEG Tokens.} Sa2VA connects SAM 2 and the LMM via the special token ``[SEG]''. The hidden states of the ``[SEG]'' token are used as a new type of prompt and fed into SAM 2's Decoder, where they are decoded into segmentation masks. The hidden states of ``[SEG]'' can be seen as a novel spatial-temporal prompt for SAM 2. SAM 2 segments the corresponding object mask in image and video based on the spatial-temporal prompt. During training, the SAM 2 decoder can be tuned to understand the spatial-temporal prompt, and gradients can be backpropagated through the ``[SEG]'' token to the LMM, allowing it to output the spatial-temporal prompt better.

\begin{algorithm}[!t]
\footnotesize
\caption{RVOS Inference Pipeline}\label{alg:refvos_inf}
\textbf{Input:} Video length $N$; Number of key frames $M$; Video frames $S_{N}$ ($X_1$, $X_2$, $X_3$,$\ldots$, $X_N$); Language description $T$;\\
\textbf{Output:} Sequence of masks $M_1$, $M_2$, $M_3$,$\ldots$, $M_N$;\\
\textbf{Run:} Sa2VA Model for RVOS;\\
Uniform sampling to extract key frames: $S_{M}$ $\gets$ $S_{N}$;\\
Visual embeddings: $E_v$ $\gets$ Encoder($S_{M}$);\\
Language embeddings: $E_l$ $\gets$ Encoder($T$);\\
Answers: $A$ $\gets$ LLM($\{E_v, E_l\}$);\\
Prompt embedding: $P_l$ $\gets$ Linear(Find($A$, '[SEG]'));\\
\For{$i = 1,2,\ldots,M$}{
    SAM 2 feature: $F_{i}$ $\gets$ Encoder($X_0$);\\
    Mask: $M_i$ $\gets$ Decoder($\{P_l, F_{i}\}$);\\
    Update Memory: $Mem$ $\gets$ Cross-Attention($\{Mem, M_i\}$);\\
}
\For{$i = M+1,M+2,\ldots,N$}{
    SAM 2 feature: $F_{i}$ $\gets$ Encoder($X_0$);\\
    Mask: $M_i$ $\gets$ Decoder($\{Mem, F_{i}\}$);\\
    Update Memory: $Mem$ $\gets$ Cross-Attention($\{Mem, M_i\}$);\\
}
\textbf{emit} $M_1$, $M_2$, $M_3$,$\ldots$, $M_N$;
\end{algorithm}

\subsection{Inference}
\label{sec:inference}
For RVOS tasks, Sa2VA designs a simple framework to achieve strong results on public benchmarks. In particular, for giving input video, it adopts a ``[SEG]'' token to generate the masks of the key frames. Then, it uses the memory encoded by the key frame features to generate the mask for the remaining frames. Sa2VA defaults to extracting the first five frames of the input video as key frames into LLM, but MeViS is a long video dataset, which results in a significant loss of video information. As described in \cref{alg:refvos_inf}, we uniformly sample the video frames as key frames to enhance the LMM's understanding of the entire video.

These key frames are fed into CLIP and flattened to visual sequential tokens for LLM processing. The LLM takes the visual and language tokens as input and uses these tokens to extract information about the video to generate the ``[SEG]'' token. In SAM 2, the prompt encoder encodes boxes or clicks to prompt embeddings for object referring. Different from SAM 2, Sa2VA use two linear layers to project the ``[SEG]'' token into the language prompt embedding, which serves as an extension of the SAM 2 prompt encoders. With the language prompt embedding, it uses the SAM 2 decoder to generate the masks of the key frames. Then, Sa2VA use the memory encoder of SAM 2 to generate a memory based on the output key-frame masks. Finally, the memory attention in SAM 2 generates the remaining masks using the memory generated from the key-frame and previous non-key-frame masks.

\subsection{Aggregation}
\label{sec:aggregation}
We find that Sa2VA does not necessarily perform better with a larger number of parameters and more sampling frames, as each configuration has its own strengths in different videos. And for some videos that cannot be accurately segmented by LMMs, the classic RVOS model may handle them very well. So we integrate the results of multiple expert models to mitigate the erroneous predictions of a single model:
\begin{equation}
    \mathcal{M} = \mathcal{F}^{fuse}\left(\mathcal{M}^{K}\right),
\end{equation}
where $\mathcal{M}^{K}$ is the $K$ sets of mask sequences output by Sa2VA models with different configurations and other RVOS models~\cite{fang2024uninext}, $\mathcal{F}^{fuse}$ denotes pixel-level binary mask voting. If there are more than $(N+1)/2$ pixels with a value equal to 1, we divide the pixel into the foreground, otherwise, it is divided into the background.

\section{Experiments}
\label{sec:exper}
\subsection{Datasets and Metrics}
\label{sec:data}
\noindent \textbf{Dataset.}
MeViS~\cite{ding2023mevis} is a newly established dataset that is targeted at motion information analysis and contains 2,006 video clips and 443k high-quality object segmentation masks, with 28,570 sentences indicating 8,171 objects in complex environments. All videos are divided into 1,662 training videos, 190 validation videos and 154 test videos.

\noindent \textbf{Evaluation Metrics.}
we employ region similarity $\mathcal{J}$ (average IoU), contour accuracy $\mathcal{F}$ (mean boundary similarity), and their average \( \mathcal{J} \)\&\( \mathcal{F} \) as our evaluation metrics.

\subsection{Implementation Details}
\label{sec:imple}
We use the trained weights provided by Sa2VA~\cite{yuan2025sa2va}, which combine SOTA LMMs models like InternVL2.5~\cite{chen2024expanding} and SAM 2~\cite{ravi2024sam}. Sa2VA is trained on four types of datasets, including image QA, video QA, image segmentation, and video segmentation datasets. For video-level referring expression segmentation, Sa2VA used 5.8K existing RVOS data from Ref-YouTubeVOS~\cite{seo2020urvos}, MeViS~\cite{ding2023mevis}, ReVOS~\cite{yan2024visa} and 37K self-built Ref-SAV~\cite{yuan2025sa2va} dataset. We conduct testing on a NVIDIA A800 GPU with 80GB of memory.

\subsection{Main Results}
\label{sec:resul}
As shown ~\cref{tab:leader}, our solution achieves 61.98 $\mathcal{J} \& \mathcal{F}$ on the MeViS test set and ranks 1st place in the 4th PVUW Challenge MeViS Track at CVPR 2025.

\begin{table}
\caption{The leaderboard of the MeViS test set.}
\setlength\tabcolsep{10pt}
  \centering
  \begin{tabular}{l|ccc}
    \toprule
    Team & \( \mathcal{J} \)\&\( \mathcal{F} \) & $\mathcal{J}$ & $\mathcal{F}$ \\
    \midrule
    \bf MVP-Lab & \bf 61.98 & \bf 58.83 & \bf 65.14\\
    ReferDINO-Plus & 60.43 & 56.79 & 64.07\\
    HarborY & 56.26 & 52.68 & 59.84\\
    Pengsong & 55.91 & 53.06 & 58.76\\
    ssam2s & 55.16 & 52.00 & 58.33\\
    strong\_kimchi & 55.02 & 51.78 & 58.27\\
  \bottomrule
\end{tabular}
\label{tab:leader}
\end{table}

\subsection{Ablation Study}
\label{sec:ablat}
To validate the effectiveness of model, we conduct simple ablation studies. As shown in ~\cref{tab:ablation}, Model is parameter quantity of Sa2VA, Sampling is whether uniform sampling, and Number refers to the number of frames sampled in the video. When sampling the first 5 frames, the improvement of Sa2VA-26B is only 0.51\% \( \mathcal{J} \)\&\( \mathcal{F} \) compared to Sa2VA-8B, indicating that the potential of the model has not been fully utilized. As shown in the third row, using uniform sampling improved by 3.58 \% \( \mathcal{J} \)\&\( \mathcal{F} \), indicating that this is crucial for correctly understanding the entire video. Increasing the sampling frame number is still effective, achieving the highest performance of 58.06\% \( \mathcal{J} \)\&\( \mathcal{F} \) at 20 frames. As shown in ~\cref{tab:uninext}, without any post-processing or semi-supervised learning, Sa2VA~\cite{yuan2025sa2va} reachs a level comparable to UNINEXT-Cutie~\cite{fang2024uninext}.

\begin{table}
\setlength\tabcolsep{5pt}
\caption{Sa2VA on the MeViS validation set.}
  \centering
  \begin{tabular}{c|c|c|ccc}
    \toprule
    Model & Sampling & Number & \( \mathcal{J} \)\&\( \mathcal{F} \) & $\mathcal{J}$ & $\mathcal{F}$ \\
    \midrule
    8B & \ding{55} & 5 & 51.60 & 48.19 & 55.01\\
    26B & \ding{55} & 5 & 52.11 & 48.35 & 55.88\\
    26B & \ding{51} & 5 & 55.69 & 51.93 & 59.44\\
    26B & \ding{51} & 10 & 57.45 & 53.86 & 61.03\\
    26B & \ding{51} & 15 & 57.86 & 54.32 & 61.40\\
    26B & \ding{51} & 20 & \bf 58.06 & 54.61 & \bf 61.52\\
    26B & \ding{51} & 25 & 57.98 & \bf 54.73 & 61.23\\
  \bottomrule
\end{tabular}
\label{tab:ablation}
\end{table}

\begin{table}
\caption{Comparison with Sa2VA and UNINEXT-Cutie on the MeViS val set.}
\setlength\tabcolsep{10pt}
  \centering
  \begin{tabular}{l|ccc}
    \toprule
    Team & \( \mathcal{J} \)\&\( \mathcal{F} \) & $\mathcal{J}$ & $\mathcal{F}$ \\
    \midrule
    SA2VA-26B~\cite{yuan2025sa2va} & 58.06 & 54.61 & \bf 61.52\\
    UNINEXT-Cutie~\cite{fang2024uninext} & \bf 58.93 & \bf 56.39 & 61.46\\
  \bottomrule
\end{tabular}
\label{tab:uninext}
\end{table}


\section{Conclusion}
\label{sec:concl}
In this work, we propose a simple and effective inference optimization method to fully unleash the potential of LMMs in referring video segmentation. Firstly, we use Sa2VA as our baseline, which is a unified LMM for dense grounded understanding of both images and videos. Secondly, we uniformly sample the video frames during the inference process to enhance the model's understanding of the entire video. Finally, we integrate the results of multiple expert models to mitigate the erroneous predictions of a single model. Our solution achieved 61.98\% $\mathcal{J} \& \mathcal{F}$ on the MeViS test set and ranked 1st place in the 4th PVUW Challenge MeViS Track at CVPR 2025.

{
    \small
    \bibliographystyle{ieeenat_fullname}
    \bibliography{main}
}

\end{document}